\documentclass{article}

\usepackage{natbib}
\usepackage{amsfonts}
\usepackage{amsmath, amssymb}
\usepackage{bm}

\usepackage{dsfont}

\usepackage{hyperref}
\usepackage{times}
\usepackage{graphicx} % more modern
\usepackage{graphics}
\usepackage{url}
\usepackage{epstopdf}
\usepackage{xcolor}
\usepackage{colortbl}
\usepackage[11pt]{moresize}
\usepackage{algorithm}
\usepackage{algpseudocode}
\usepackage[justification=centering]{caption}
\usepackage{wrapfig}

\DeclareMathOperator{\E}{\mathbb{E}}
\newcommand{\kp}{{k p}}

\usepackage[preprint]{neurips_2023}

\usepackage[utf8]{inputenc} % allow utf-8 input
\usepackage[T1]{fontenc}    % use 8-bit T1 fonts
\usepackage{hyperref}       % hyperlinks
\usepackage{url}            % simple URL typesetting
\usepackage{booktabs}       % professional-quality tables
\usepackage{amsfonts}       % blackboard math symbols
\usepackage{nicefrac}       % compact symbols for 1/2, etc.
\usepackage{microtype}      % microtypography
\usepackage{caption}
\usepackage{subcaption}

\title{Spotlight Attention: Robust Object-Centric Learning With a Spatial Locality Prior}

\author{%
  Ayush Chakravarthy \\
  Mila, Université de Montréal \\
  \texttt{ayush.k.chakravarthy@gmail.com} \\
  \And
  Trang Nguyen \\
  Mila, Université de Montréal \\
  \AND
  Anirudh Goyal \\
  Google DeepMind \\
  \And
  Yoshua Bengio \\
  Mila, Université de Montréal  \\
  \And
  Michael C. Mozer \\
  Google Research, Brain Team \\
}

\begin{document}

\maketitle

\begin{abstract}
The aim of object-centric vision is to construct an explicit representation of the objects in a scene. This representation is obtained via a set of interchangeable modules called \emph{slots} or \emph{object files} that compete for local patches of  an image. The competition has a weak inductive bias to preserve spatial continuity; consequently, one slot may claim patches scattered diffusely throughout the image.  In contrast, the inductive bias of human vision is strong, to the degree that attention has classically been described with a spotlight metaphor. We incorporate a spatial-locality prior into state-of-the-art object-centric vision models and obtain significant improvements in segmenting objects in both synthetic and real-world datasets. Similar to human visual attention, the combination of image content and spatial constraints yield robust unsupervised object-centric learning, including less sensitivity to model hyperparameters.

\end{abstract}

\section{Introduction}
\label{sec1}
Learning about objects and their interactions is a cornerstone of human cognition \citep{https://doi.org/10.1111/j.1467-7687.2007.00569.x}. Understanding the nature of objects and their properties is necessary to achieve symbol-like mental representations \citep{Whitehead1928} and systematicity of reasoning \citep{FodorPylyshyn88a}. Although language has a natural tokenization that supports systematicity \citep{chakravarthy-etal-2022-systematicity}, progress in visual reasoning in Artificial Intelligence hinges on tokenizing visual input. Visual tokenization corresponds to the problem of \textit{object-centric representation  learning}---learning to partition images into a set of discrete slots in an unsupervised manner. The desiderata for these slots are that they induce a bijection with the objects in an image and are interchangeable. 

With no or limited supervision, object-centric representation learning is very difficult \citep{Greff2020OnTB} and requires appropriate forms of inductive bias \citep{Scholkopfetal21, ke2021systematic, goyal2022inductive}. The biases explored for building models with object-centric representation have been based on instance based segmentation \citep{Greff2017NEM, Greff2019MultiObjectRL, Locatello2020}, sequential object extraction \citep{pmlr-v37-gregor15, burgess2019monet, Engelcke2021GENESISV2IU, Goyaletal21RIM},  invariance \citep{Crawford2019autoyolo, Lin2020SPACEUO, JiangJanghorbaniDeMeloAhn2020SCALOR, Biza2023InvariantSA}, type-token distinction \citep{bao2022discovering, bao2023object, Goyal2020} and the sparsity of interactions among different slots \citep{alias2021neural, goyal2021coordination}.

However, none these methods leverage a bias that is considered fundamental to visual attention in the psychological literature: the preference for spatial continuity of an attended region.  Traditionally, attention was considered to be a \emph{spotlight} on a region of interest in the image \citep{Posner}.  The spotlight metaphor was extended to be a zoom lens \citep{LaBerge}, allowing for large or small regions, but nonetheless the regions needed to be convex. The metaphor was further extended to allow the selected region to blanket around a shape \citep{Mozer1988}.  In the psychological literature, even the notion of \emph{object-based attention} \citep{Duncan} is spatial in nature  \citep{Vecera}, although it allows noncontiguous features of an object to be selected together
e.g., if two ends of an occluded object are visible \citep{ZemelBehrmannMozer}.

Although current techniques in object-centric representation learning make available a positional encoding to guide the mapping between image patches and slots, the slot extraction process itself has no explicit pressure to choose patches in a spatially contiguous manner. Consequently, one slot may claim patches scattered diffusely throughout the image. In this paper, we introduce an inductive bias in the form of a \emph{spatial locality prior} (\emph{SLP}) that encourages slots to select spatially contiguous patches in the input image. 

\textbf{Summary.} We present an algorithm that can be incorporated into models for object-centric representation learning that biases slot assignments based on spatial locality. We show consistent improvements in the quality of object representations for three object-centric architectures, eight distinct data sets, both synthetic and natural, and multiple different performance measures that have been used in the literature.  We show that the SLP also makes the baseline models more robust to hyperparameter selection and supports robust out-of-distribution generalization.

\iffalse
\begin{itemize}
    \item We present an algorithm that can be incorporated into can be incorporated into models for object-centric representation learning  (e.g., Slot Attention, BO-QSA, DINOSAUR) that biases slot assignments based on spatial locality.
    \item We show consistent improvements in the quality of object representations for three object-centric architectures, eight distinct data sets, both synthetic and natural, and four different performance measures that have been used in the literature.
    \item We show that the SLP also makes the baseline models more robust to hyperparameter selection and supports robust out-of-distribution generalization.
\end{itemize}
\fi

\section{Background}
\textit{Unsupervised Object-Centric Learning.} Our work falls into the line of research in machine vision known as unsupervised object-centric representation learning \citep{eslami2016, Greff2017NEM, Greff2019MultiObjectRL, burgess2019monet, Goyal2020, Lin2020SPACEUO, Locatello2020, Engelcke2021GENESISV2IU, Singh2022SLATE, jia2022unsupervised}. Broadly, this work has the objective of mapping image elements to a low-dimensional set of objects, or \emph{slots}, where the goal is for grouped image elements to be semantically similar. As `semantic similarity' is an ambiguous objective, several additional sources of information have been explored including video sequences \citep{JiangJanghorbaniDeMeloAhn2020SCALOR, weis2022, singh2022simple, traub2023learning} and optical flow \citep{kipf2022savi, elsayed2022savipp, bao2022discovering, bao2023object}. However, as pointed out by \cite{yang2022}, such methods suffer from problems in scaling to large real-world datasets. Toward alleviating this problem, recent work has moved from CNNs to Transformer-based models which have greater expressivity \citep{Singh2022SLATE, Seitzer2023}. Nonetheless, the basic mechanism of slot assignment in Slot Attention \citep{Locatello2020} and its variants \citep{Chang2022ObjectRA, jia2022unsupervised}  has proven 
a critical building block to unsupervised object-centric learning.

\textit{Spatially-biased OCL.}
% Spatially biased OCL?
% OCL with a spatial prior?
The key novelty of our work is a spatial-proximity-based mechanism that modulates the slot-pixel assignment. One might argue that because current methods for object centric representation learning includes explicit positional encodings in its input \citep{Locatello2020, Goyal2020, alias2021neural}, the existing methods have sufficient information to discover the principle of spatial coherence. However, the results we present clearly indicate otherwise, or at least that spatial constraints have yet to be fully exploited. Other work has incorporated spatial information into models, such as using shape priors for weak supervision \citep{Elich2020WeaklySL}, spatial position as a means to bootstrap learning \citep{Kim2023ShepherdingST}, and incorporating spatial symmetries \citep{Biza2023InvariantSA}. Furthermore, in the context of video, SaVi \citep{kipf2022savi} and SaVi++ \citep{elsayed2022savipp} use ground-truth spatial information such as center of mass and bounding boxes extracted from the first frame of the video. However, our method is the first to use spatial constraints to steer the slot-pixel assignment with no supervision.

\section{Method}
\subsection{Augmenting Key-Query Match with Spatial Bias}
The input image is first pre-processed by an \emph{image encoder}. This processing preserves the image topography, yielding an embedding at each image patch $p$, which corresponds to an $(x,y)$ position in a coarse grid over the image. (The embedding is intended to encode high-level visual features, and thus the grid of patches is coarser than the input dimensions in pixels.) The embedding at each patch $p$ is mapped to a key, $\bm{\kappa}_p$, which is matched to a query from each slot $k$, denoted $\bm{q}_k$. The evidence supporting a key-query match is
\[
\gamma_\kp = \frac{\bm{q}_k^\mathrm{T} \bm{\kappa}_p}{\sqrt{d}},
\]
where $d$ is the dimensionality of the query and key vectors.

In various different methods of object centric representation learning \citep{Goyal2020, Locatello2020, alias2021neural}, slots compete for each grid position via a softmax renormalization of the match scores $\gamma_\kp$. Here, we introduce a \emph{Spatial Locality Prior} to modulate the competition among slots. This prior takes the form of an additive term, $\alpha_\kp$, in the softmax:
\[
 s_\kp = \mathrm{softmax}_{k}\big(\gamma_\kp + \alpha_\kp\big),
\]
where $s_\kp$ is the affinity between position $p$ and slot $k$, and $\mathrm{softmax}_k$ denotes the $k$'th element of the softmax vector. We use $\bm{s}_k \equiv \{ s_{k} \}$ to denote the distribution of activation (or attention) over positions for a given slot $k$. The $\bm{\alpha}$ matrix is used to bias this distribution by spatial locality.

\subsection{Encouraging Spatial Locality}

Conditioned on an input, the $\bm{\alpha}$ matrix is determined by a constraint satisfaction process (CSP) that encourages a roughly spotlight-like distribution of activation over positions in $\bm{s}_k$ for each slot $k$. Additionally, the CSP discourages overlap in the spotlights of any pair of slots. The CSP converges on $\bm{\alpha}$ by iterative gradient-descent steps in a loss that penalizes spotlights that are non-compact and overlapping.

The spotlight associated with each slot $k$ is characterized by its  center, $\bm{m}_k$, and isotropic spread, $v_k$, defined to be the central tendency and variance of $\bm{s}_k$:
\begin{equation} \label{eqn:dist}
    \bm{m}_k = \frac{\sum_p s_\kp \bm{p}}{\sum_p s_\kp} \text{~~~~and~~~~}
v_k = \frac{\sum_p s_\kp |\bm{p}-\bm{m}_k|^2}{\sum_p s_\kp}
\end{equation}

Note that while $\sum_k s_\kp=1$ due to the softmax, the sum over positions is not normalized. 

The CSP's loss consists of two terms. First, a penalty is imposed to the degree that each pair of slots  fail to have spatially distinct attentional profiles, as characterized by a distance measure summed over slot pairs:
\begin{equation}  \label{eqn:loss_1}
  \mathcal{L}_\mathrm{distinct} = \sum_{k,k'>k}  \exp \left( - 
     \frac{|\bm{m}_k - \bm{m}_{k'}|^2}{v_k + v_{k'}} 
     \right)  
\end{equation}

This loss is designed to push apart the slot means (the numerator term) relative to the intra-slot variance (the denominator term).
Second, to prevent degenerate solutions in which attention collapses to a point, we impose a penalty on the Froebenius
norm of $\bm{\alpha}$:
\begin{equation} \label{eqn:loss_2}
  \mathcal{L}_\mathrm{norm} = \sum_{k,p} \alpha_\kp^2  
\end{equation}

The overall loss $\mathcal{L} = \mathcal{L}_\mathrm{distinct} + \lambda \mathcal{L}_\mathrm{norm}$ is minimized with respect to $\bm{\alpha}$ from an
initial state $\bm{\alpha}^0$, which we discuss next.

\begin{algorithm}[t]
\caption{\textbf{Spatial Locality Prior.} The algorithm takes image embedding features $ Z \in \mathbb{R}^{N \times C}$; the number of slots, $K$; the number of slot-update iterations, $T_\textit{slot}$; the number of spatial-update iterations, $T_\textit{spat}$; and the projection dimensionality, $d$. The learned model parameters are: the learned projections $q, k, v$ each with dimensionality $d$, the alpha initialization $\bm{\alpha}^0 \in \mathbb{R}^{K \times N}$; the \textbf{GRU} and the \textbf{MLP} layers; and a Gaussian mean and variance $\mu, \sigma \in \mathbb{R}^d$. }\label{alg:ssa}
\begin{algorithmic}
%\State $S = \textrm{\textbf{Tensor}}(K, d)$
\State $S \sim \mathcal{N}(\mu, \sigma)$
\State $Z = \textrm{\textbf{LayerNorm}}(Z)$
\State $\bm{\alpha} = \bm{\alpha}^0 \  / \  ||\bm{\alpha}^0||_2$
\For{$i = 1 \dots T_\textit{slot}$}
    \State $S = \textrm{\textbf{LayerNorm}}(S)$
    \State $L = \frac{1}{\sqrt{d}}q(S) \cdot k(Z)^\textsc{T}$
    \For{$j = 1 \dots T_\textit{spat}$}
        \If{$j = T_\textit{spat} - 1$}
            \State $\bm{\alpha} = \textrm{\textbf{StopGradient}}(\bm{\alpha}) + \bm{\alpha}^0 - \textrm{\textbf{StopGradient}}(\bm{\alpha}^0)$
        \EndIf
        \State $A = \textrm{\textbf{Softmax}}(L + \bm{\alpha}, \textrm{axis}=\textrm{"slots"})$
        \State $m, v = \textrm{\textbf{ComputeDistribution}}(Z, A)$ \Comment{Compute Equation \ref{eqn:dist}}
        \State $l = \textrm{\textbf{GetLoss}}(\bm{\alpha}, m, v)$ \Comment{Compute Equations \ref{eqn:loss_1} and \ref{eqn:loss_2}}
        \State $\bm{\alpha} = \bm{\alpha} - \bm{\alpha}_{\textrm{lr}} \cdot \frac{\partial l}{\partial \bm{\alpha}}$
    \EndFor
    \State $A = \textrm{\textbf{Softmax}}(L + \bm{\alpha}, \textrm{axis}=\textrm{"slots"})$
    \State $A = A \  / \  \textrm{\textbf{sum}}(A, \textrm{axis}=\textrm{"embeddings"})$
    \State $U = A \cdot v(Z)$
    \For{$n = 1 \dots N$ \textbf{in parallel}}
        \State $S_n = \textrm{\textbf{GRU}}(S_n, U_n)$
        \State $S_n \mathrel{+}= \textrm{\textbf{MLP}}(\textrm{\textbf{LayerNorm}}(S_n))$
    \EndFor
\EndFor
\State \Return $S$
\end{algorithmic}
\end{algorithm}

\subsection{Learning Initial State Through Bilevel Optimization}

To break symmetry, it is vital to learn an initial state $\bm{\alpha}^0$ which partitions the image by dispersing initial slot means across the image.  We take inspiration from \cite{jia2022unsupervised} and perform meta-learning to determine $\bm{\alpha}^0$. On each training trial, after $j$ steps of the CSP optimization, we obtain an approximately optimal attentional bias, let's call it $\bm{\alpha}^*$. We detach $\bm{\alpha}^*$ and optimize for $\bm{\alpha}^0$ on the last CSP step. We use the straight-through estimator \citep{Bengio2013EstimatingOP, Oord2017NeuralDR} to additionally propagate gradients into $\bm{\alpha}^0$. Through this design, we are able to learn generalized dataset-wide statistics about $\E[\bm{m}_k]$ and $\E[v_k]$ for each slot $k$. Note that this procedure breaks slot symmetry because $\bm{\alpha}^0$
assigns slots to default regions of the image. 

%in which the slot means and dispersed through the scene and the slot variances are small. 
%We take inspiration from \cite{jia2022unsupervised} and initialize $\bm{\alpha}^0$ to be a learnable query. 
%After $j$ steps of $\bm{\alpha}$ optimization to obtain an approximation of $\bm{\alpha}^*$, the approximate optimal is detached and then passed for one last $\bm{\alpha}$ optimization step. We use STE \citep{Bengio2013EstimatingOP, Oord2017NeuralDR} to additionally propagate gradients into $\bm{\alpha}^0$.

\section{Experiments}

In this section, we evaluate the benefit of incorporating the spatial-locality prior into Slot Attention, as well as into two recent object-centric methods that build upon Slot Attention to achieve state-of-the-art performance, BO-QSA \citep{jia2022unsupervised} and DINOSAUR \citep{Seitzer2023}. We find that for all three models, across eight diverse datasets, the spatial-locality prior boosts performance. 
We refer to the base models (Slot Attention, BO-QSA, and DINOSAUR) augmented with the spatial
locality prior (hereafter, \emph{SLP}) by adding the modifier `+SLP' to the name. All comparisons we report use a given base model with and without SLP for highly controlled experimentation.

Here we include details about our datasets, architectural decisions, and evaluation methods; but hyperparameters and other simulation details are presented in the Supplementary Materials. All experiments were run on a single 50GB Quatro RTX 8000 GPU. 
%In this section we aim to address the following questions through empirical results:
%\begin{enumerate}
%%    \item How does the proposed method compare to various other methods for object-centric representation learning?
%    \item Can the proposed method be used to augment various other methods for object-centric representaiton learning?
%\end{enumerate}

\begin{table}[t]
\caption{Foreground ARI (\%) Segmentation Accuracy \\ (mean ~± 1 SEM across 3 replications of each simulation)}
\centering
\begin{subtable}[t]{0.99\textwidth}
\caption{Results for Vanilla Slot Attention augmented with SLP }
\centering
\resizebox{\columnwidth}{!}{%
\begin{tabular}{ lccccc} 
\toprule
Method & \textbf{ObjectsRoom} & \textbf{MultidSprites} & \textbf{ShapeStacks} & \textbf{ClevrTex} \\
\midrule
MONet \citep{burgess2019monet} &
0.54 \textcolor{gray}{\ssmall ~± 0.05} & 
0.89 \textcolor{gray}{\ssmall ~± 0.05} &
0.70 \textcolor{gray}{\ssmall ~± 0.11} &
0.19 \textcolor{gray}{\ssmall ~± 0.05}

\\

GENESIS-V2 \citep{Engelcke2021GENESISV2IU} &
0.86 \textcolor{gray}{\ssmall ~± 0.05} &
0.52 \textcolor{gray}{\ssmall ~± 0.15} & 
0.81 \textcolor{gray}{\ssmall ~± 0.05} &
0.31 \textcolor{gray}{\ssmall ~± 0.20}
\\

Slot Attention  \citep{Locatello2020} & 
0.86 \textcolor{gray}{\ssmall ~± 0.14} & % not done
0.91 \textcolor{gray}{\ssmall ~± 0.10}  & % not done
0.80 \textcolor{gray}{\ssmall ~± 0.08} &
0.62 \textcolor{gray}{\ssmall ~± 0.08} \\ % not done

\midrule

Slot Attention + SLP  & 
\cellcolor{blue!15}0.87 \textcolor{gray}{\ssmall ~± 0.05}  & % done
\cellcolor{blue!15}0.94 \textcolor{gray}{\ssmall ~± 0.05}  & %done
\cellcolor{blue!15}0.83 \textcolor{gray}{\ssmall ~± 0.05} &
\cellcolor{blue!15}0.71 \textcolor{gray}{\ssmall ~± 0.05}\\% not done
\bottomrule
\end{tabular}%
}
\label{tab:ssasynth}
% \vspace{-0.1cm}
\end{subtable}
\hfill
\begin{subtable}[t]{0.99\textwidth}
\caption{Varying hyperparameters (number of slot-update iterations and \\spatial-update iterations) for ClevrTex}
\centering
\resizebox{\columnwidth}{!}{%
\begin{tabular}{ lcccccc} 
\toprule
Method & {3 slot iterations} & {5 slot iterations} & {10 slot iterations} \\
\midrule
Slot Attention \citep{Locatello2020} & 
0.45 \textcolor{gray}{{\ssmall ~± 0.23}} &  %3
0.35 \textcolor{gray}{{\ssmall ~± 0.20}} & %5
0.34 \textcolor{gray}{{\ssmall ~± 0.16}} \\ %10

\midrule

Slot Attention + SLP (1 spatial iteration) & 
0.54 \textcolor{gray}{\ssmall ~± 0.05} &   % 3
0.61 \textcolor{gray}{\ssmall ~± 0.05} & % 5
0.60 \textcolor{gray}{\ssmall ~± 0.14} \\ % 10

Slot Attention + SLP (5 spatial iterations) & 
\cellcolor{blue!15}0.64 \textcolor{gray}{\ssmall ~± 0.10} &   % 3
0.60 \textcolor{gray}{\ssmall ~± 0.08} & % 5
0.63 \textcolor{gray}{\ssmall ~± 0.11} \\ % 1

Slot Attention + SLP  (10 spatial iterations) & 
\cellcolor{blue!15}0.65 \textcolor{gray}{\ssmall ~± 0.08} &   % 3
0.63 \textcolor{gray}{\ssmall ~± 0.08} & % 5
0.62 \textcolor{gray}{\ssmall ~± 0.05} \\ % 10

\bottomrule
\end{tabular}%
}

\label{tab:ali-ablate}
%\vspace{-0.3cm}
\end{subtable}
\end{table}

\begin{figure}
    \centering
    \begin{subfigure}[b]{\linewidth}  
    \centering
    \includegraphics[width=\linewidth]{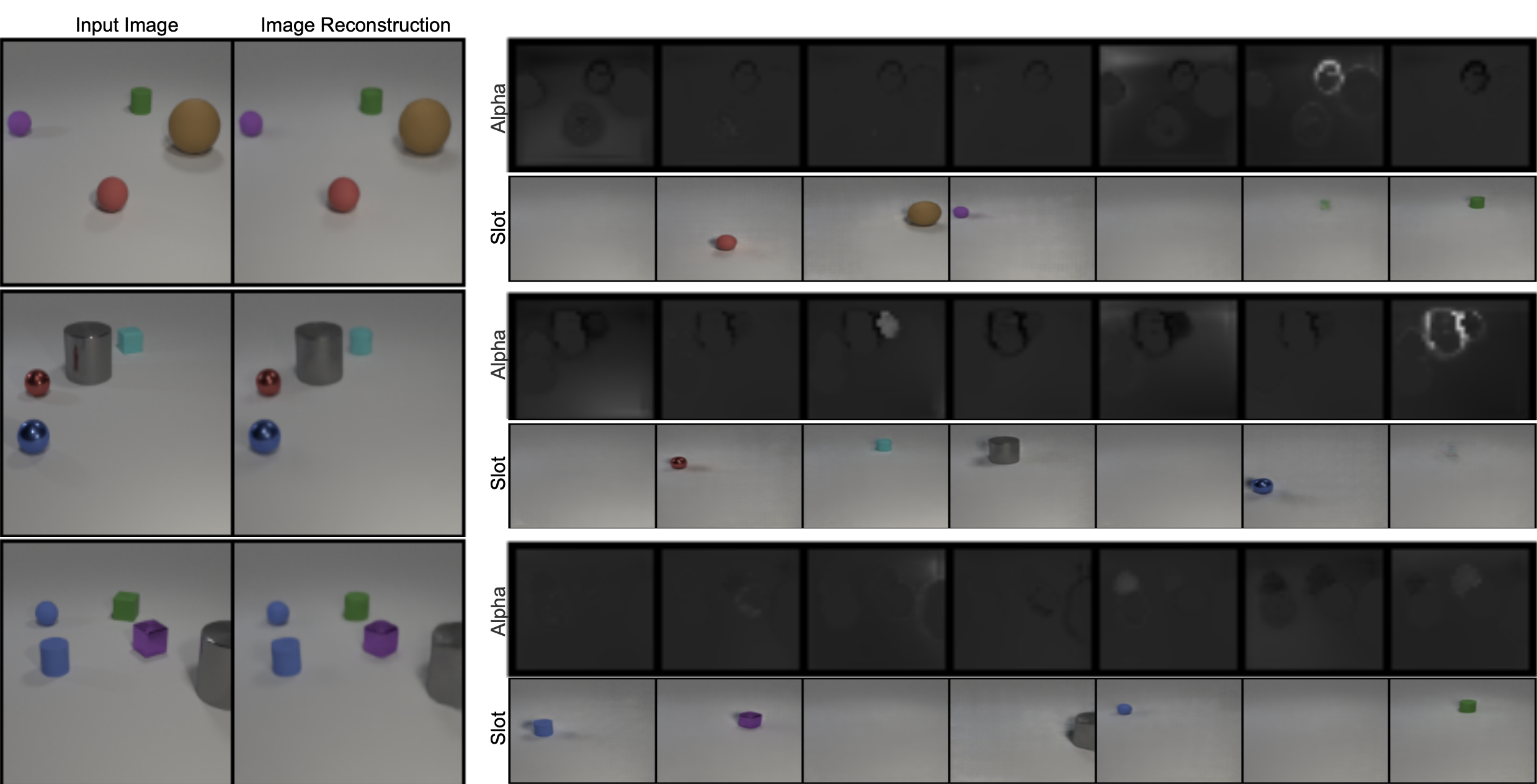}
    \caption{Clevr6 examples: Visualizations for Image Reconstructions, Slot Extractions, and Alphas.} %of three input images (first column), reconstructions (second column), slot-wise $\bm{\alpha}_k$ spatial attentional biases (darker panel to the right of each image), and slot representation (colored panel to the right of each image)} 
    \label{fig:alpha}
    \end{subfigure}
    \begin{subfigure}[b]{\linewidth}
    \centering
    \includegraphics[width=\linewidth]{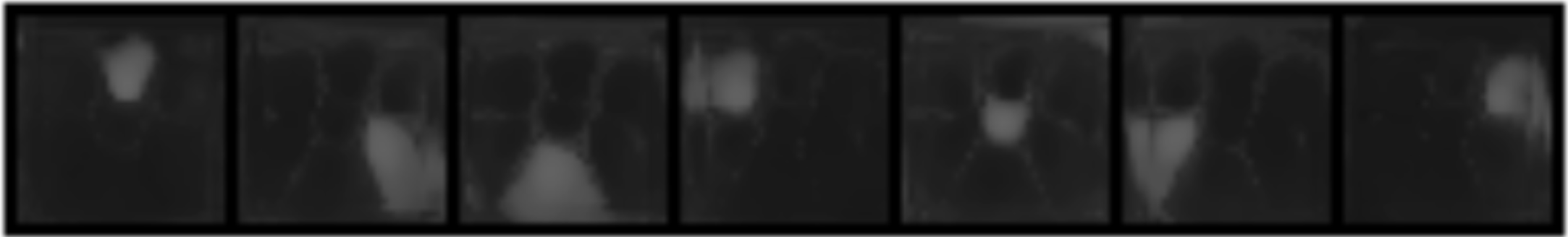}
    \caption{Visualization of meta-learned $\bm{\alpha}^0$ on Clevr6}
    \label{fig:alpha_init}
    \end{subfigure}
    %\vspace{-0.5cm}
    \caption{Results from Slot Attention + SLP}
\end{figure}

\subsection{Object Discovery in Synthetic Images}

For synthetic datasets, we focus on the task of Object Discovery \citep{burgess2019monet}, which is to produce a set of masks that cover each of the objects that appear in the image. We first isolate the effect of SLP through experimenting with vanilla Slot Attention \citep{Locatello2020} on CLEVR6 \citep{Johnson2016CLEVRAD}, ObjectsRoom \citep{multiobjectdatasets19}, MultidSprites \citep{burgess2019monet}, ShapeStacks \citep{ShapeStacks}, and ClevrTex \citep{Karazija2021ClevrTexAT}. To show that SLP works for other variants of Slot Attention, we examine BO-QSA \citep{jia2022unsupervised} on ShapeStacks, ObjectsRoom, and ClevrTex datasets. We primarily focus on Foreground-ARI (FG-ARI) \citep{Hubert1985ComparingP} as our dependent measure of performance. We then show results with DINOSAUR \citep{Seitzer2023} on the MoVi-C and MoVi-E datasets \citep{Greff_2022_CVPR}, where we evaluate using FG-ARI and mean-best-overlap (mBO) \citep{PontTuset2015MultiscaleCG}.

\subsubsection{Vanilla Slot Attention} \label{sssec:sa}

\begin{table}
%\begin{wraptable}{r}{6cm}
\vspace{-0.09cm}
%\tiny
  \centering
  \caption{FG-ARI (\%) Accuracy \\ (mean ~± 1 SEM across 3 runs) \\ 
  Varying number of slots for ClevrTex}
  \begin{tabular}{lcc}
    \toprule
    Model & {7 Slots} & {11 Slots} \\
    \midrule
    Slot Attention & 
    0.45 \textcolor{gray}{\tiny ~± 0.23} &
    0.62 \textcolor{gray}{\tiny ~± 0.08}    
    \\
    Slot Attention + SLP & 
    \cellcolor{blue!15}0.65 \textcolor{gray}{\tiny ~± 0.08} &
    \cellcolor{blue!15}0.71 \textcolor{gray}{\tiny ~± 0.05} \\
    \bottomrule
  \end{tabular}
  \label{tab:sa_slot_comp}
  \vspace{-0.2cm}
%\end{wraptable}
\end{table}

\textit{Methodology.} In order to isolate the effect of SLP, we setup Slot Attention as described in \citet{Locatello2020}, with a Mixture Decoder from \citet{Watters2019SpatialBD} and a 4-layer CNN encoder, on the object-discovery task. SLP is integrated into Slot Attention as
described in Algorithm \ref{alg:ssa}. Slots are initialized with a learned Gaussian mean and variance.

\textit{Results.}
Table \ref{tab:ssasynth} presents results for object discovery on synthetic datasets. Slot Attention + SLP yields improvements on MultidSprites, ShapeStacks, and, most notably, on ClevrTex. We obtain an almost 10\% improvement on ClevrTex; given challenging views with diverse textures and complicated lighting effects, Slot Attention itself is not able to effectively segment the objects. SLP does not completely solve the task, but it is a significant step forward. Further, in Table \ref{tab:sa_slot_comp}, we show that SLP allows an underparametrized Slot Attention with 7 slots to match the 11 slot baseline. Additionally, the improvement on the 7 slot  experiment is robust as a two-tailed t-test yielded $t(3)=3.61, p=.02$.

Another limitation of vanilla Slot Attention is its fragility as the number of slot-update iterations across training and evaluation phase. In Table \ref{tab:ali-ablate}, we show that SLP effectively solves this problem. In this Table, we manipulate both the number of slot-update iterations ($T_\textit{slot}$) and the number of spatial-update iterations ($T_\textit{spat}$)---the columns and rows of the table, respectively. As the number of slot-update iterations increases, performance of Slot Attention drops but Slot Attention + SLP  is not systematically affected. The Table also indicates that a single SLP iteration---minimal additional computation---provides a sufficient bias to improve FG-ARI.

The first and second columns of Figure~\ref{fig:alpha} show three sample Clevr6 images and their reconstruction, respectively. To the right of each image pair is the $\bm{\alpha}_k$ spatial-attention biases for each of the $k \in \{1...7\}$ slots (the dark images), and the the post-competition masked slot representations for each slot (color images). Note that the spatial-attention biases do not reflect the final selection, but rather the fact that SLP is encouraging slots to claim patches near already claimed patches. For some objects, especially in cases where the object seems to be distant, the spatial bias seems important, but not for all objects in these simple images.

Figure~\ref{fig:alpha_init} visualizes the meta-learned initial spatial bias distribution, $\bm{\alpha}^0$, for the 7 slots. The initialization essentially carves up the image, ignoring regions of the image that never contain objects. Note that these initial biases do not determine the final slot assignments, as one can see by inspection of the spatial distribution of objects claimed by a given  slot in Figure~\ref{fig:alpha}.

\begin{figure}[bt]
    \centering
    \includegraphics[width=\linewidth]{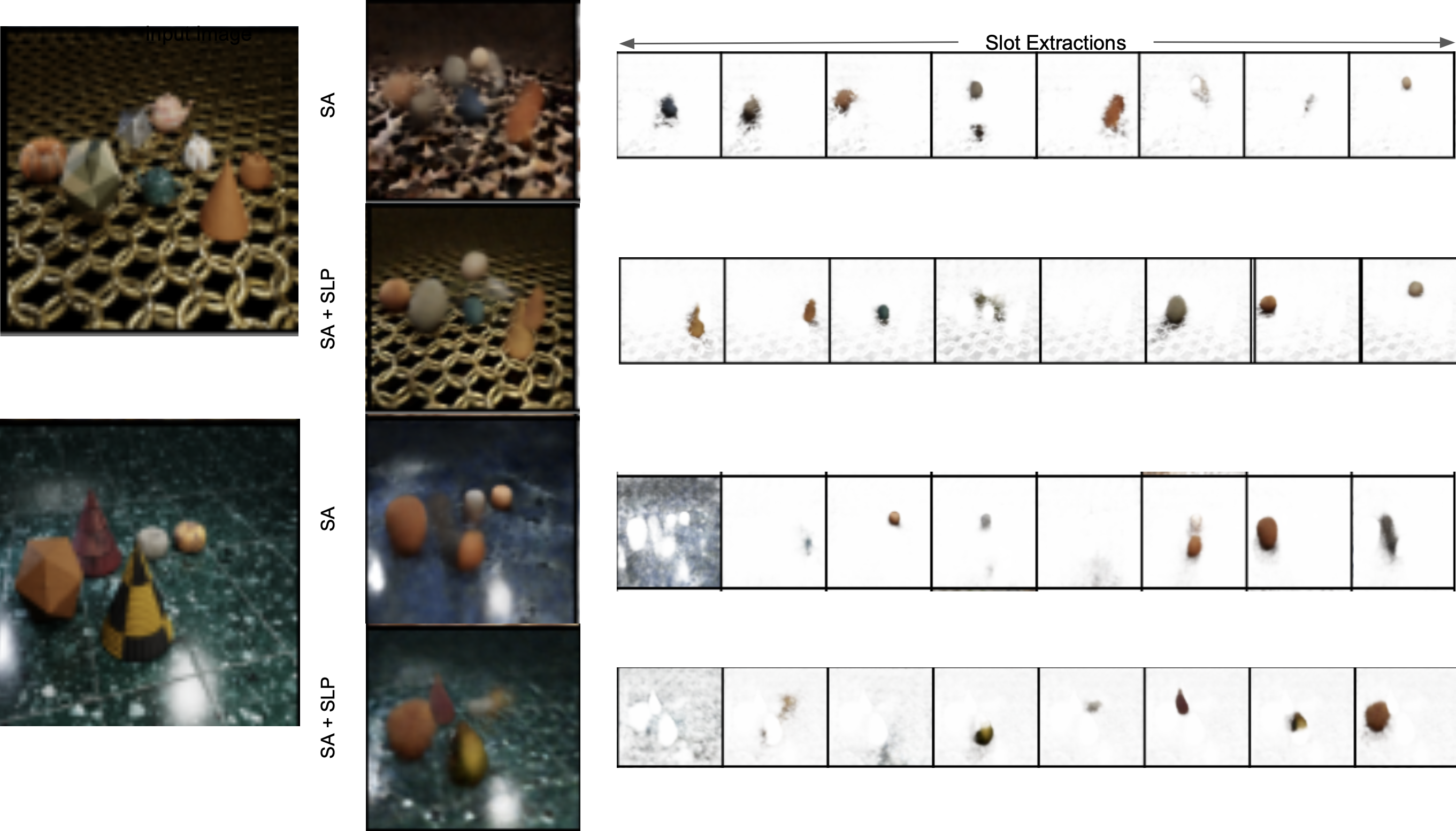}
    \caption{ClevrTex examples: Visualizing Image Reconstructions and Slot Extractions for Slot Attention (SA) and Slot Attention + SLP (SA + SLP)}
    \label{fig:savsspotlight}
\end{figure}

Figure \ref{fig:savsspotlight}
presents ClevrTex examples comparing Slot Attention and
Slot Attention (SA) + SLP via image reconstructions and slot extractions.
SA + SLP's image reconstruction is far more accurate as compared to that of Slot Attention, with the clearest difference being the quality of the background scene and the overall image sharpness. This can be explained by the observation that \textit{empty} slots of SA + SLP do a much better job of representing the background pattern. We also note that the slot decompositions are effectively localized with
SA + SLP, even though the precise boundaries of the object are not sharp.

%\begin{figure}[t]
%    \centering
%    \begin{subfigure}[b]{\linewidth}
%        \centering
%        \includegraphics[width=\linewidth]{figs/sa_vs_spotlight_recons.png}
%        \caption{Image reconstructions for a ClevrTex image. The leftmost image is ground truth, followed by the Slot Attention reconstruction, followed by the Slot Attention + SLP reconstruction}
%    \label{fig:savspotrecons}
%  \end{subfigure}
%\begin{subfigure}[b]{\linewidth}
%        \centering
% \includegraphics[width=\linewidth]{figs/sa_vs_spotlight_slotrep.png}
%        \caption{Slot decompositions for image in Figure~\ref{fig:savspotrecons} with Slot Attention (row 1) and Slot Attention + SLP (row 2)}
% \label{fig:savspotslotrep}
%   \end{subfigure}
%  \caption{Image reconstructions and decompositions with Slot Attention and Slot Attention + SLP}
%   \label{fig:savspotlight}
%\end{figure}

\subsubsection{BO-QSA} \label{ssec:bqsa}
\textit{Methodology.} Here, the setup is identical to that in Section \ref{sssec:sa}, the  only difference being that the slots are set to learnable queries which are themselves learned through gradient updates via the straight-through estimator \citep{Bengio2013EstimatingOP}.

\textit{Results.}
Table \ref{tab:boqsasynth} presents FG-ARI scores for BO-QSA. Consistent with our previous results (\ref{sssec:sa}), we observe the largest improvements on ClevrTex, while also observing statistically reliable improvements on ShapeStacks and ObjectsRoom. %We also corroborate the finding from the previous set of experiments that more spatial iterations are not necessary but they are innocuous.

\begin{table} [t]
\caption{Foreground ARI (\%) Segmentation Accuracy \\ (mean ~± 1 SEM across 3 replications of each simulation)\\
BO-QSA \citep{jia2022unsupervised} augmented with SLP}
\centering

\begin{tabular}{ lcccccc} 
\toprule
Method & \textbf{ShapeStacks} &\textbf{ObjectsRoom} & \textbf{ClevrTex} \\
\midrule
MONet \citep{burgess2019monet} & 
0.70 \textcolor{gray}{{\ssmall ~± 0.11}} &  %ss
0.54 \textcolor{gray}{{\ssmall ~± 0.05}} & %or
0.19 \textcolor{gray}{{\ssmall ~± 0.05}} \\

GENESIS-V2 \citep{Engelcke2021GENESISV2IU} & 
0.81 \textcolor{gray}{{\ssmall ~± 0.05}} &  %ss
0.86 \textcolor{gray}{{\ssmall ~± 0.05}} & %or
0.31 \textcolor{gray}{{\ssmall ~± 0.20}} \\

SLATE \citep{Singh2022SLATE} & 
0.65 \textcolor{gray}{{\ssmall ~± 0.10}} &  %ss
0.57 \textcolor{gray}{{\ssmall ~± 0.10}} & %or
0.73 \textcolor{gray}{{\ssmall ~± 0.05}} \\

I-SA \citep{Chang2022ObjectRA} &
0.90 \textcolor{gray}{{\ssmall ~± 0.08}} &
0.85 \textcolor{gray}{{\ssmall ~± 0.05}} &
0.78 \textcolor{gray}{{\ssmall ~± 0.10}} \\

BO-QSA \citep{jia2022unsupervised} & 
0.93 \textcolor{gray}{{\ssmall ~± 0.05}} &  %s
0.87 \textcolor{gray}{{\ssmall ~± 0.05}} & %or
0.80 \textcolor{gray}{{\ssmall ~± 0.08}} \\ %ct

\midrule

BO-QSA + SLP  & 
\cellcolor{blue!15}0.95 \textcolor{gray}{\ssmall ~± 0.08} &   % ss
\cellcolor{blue!15}0.93 \textcolor{gray}{\ssmall ~± 0.05} & % or
\cellcolor{blue!15}0.87 \textcolor{gray}{\ssmall ~± 0.05} \\ % ct

\bottomrule
\end{tabular} 
\label{tab:boqsasynth}
% \vspace{-0.1cm}
\end{table}

\subsubsection{DINOSAUR}
\label{sssec:dinosaur}
\begin{table}
%\begin{wraptable}{r}{8.5cm}
%\tiny
    %\vspace{-0.5cm}
    \centering
    \caption{FG-ARI and mBO measures of object discovery\\ on MoVi-C and MoVi-E with DINOSAUR \citep{Seitzer2023} (mean ~± 1 SEM across 3 runs)\\}

    \begin{tabular}{ lcccc} 
    \toprule
    Method & \multicolumn{2}{c}{\textbf{MoVi-C}} & \multicolumn{2}{c}{\textbf{MoVi-E}} \\
\cmidrule(r){2-3}   \cmidrule(r){4-5}
    & FG-ARI & mBO & FG-ARI & mBO \\
    \midrule
    DINOSAUR (ViT-B/8) &
    68.9 \textcolor{gray}{\tiny ~± 0.3} &
    38.0 \textcolor{gray}{\tiny ~± 0.2} &
    65.1 \textcolor{gray}{\tiny ~± 0.6} &
    33.5 \textcolor{gray}{\tiny ~± 0.1} \\

    DINOSAUR + SLP & 
    \cellcolor{blue!15}72.8 \textcolor{gray}{\tiny ~± 0.6} &
    \cellcolor{blue!15}41.5 \textcolor{gray}{\tiny ~± 0.3} &
    \cellcolor{blue!15}70.4 \textcolor{gray}{\tiny ~± 0.4} &
    \cellcolor{blue!15}35.9 \textcolor{gray}{\tiny ~± 0.2} \\
    \bottomrule
    \end{tabular} 
    \label{tab:dinosaursynth}
    %\vspace{0.1cm}
%\end{wraptable}
\end{table}

\textit{Methodology.} We closely follow the setup of DINOSAUR \citep{Seitzer2023}, using a frozen pre-trained ViT encoder \citep{vit}, Slot Attention with $11$ slots for MoVi-C and $24$ slots for MoVi-E, a Transformer Decoder similar to SLATE \citep{Singh2022SLATE} and feature reconstruction loss as the primary learning signal instead of image reconstruction loss.

\textit{Results.} Table~\ref{tab:dinosaursynth} compares measures of object discovery for MoVi-C and MoVi-E datasets on DINOSAUR and our augmented variant with SLP. On both data sets and two performance measures---FG-ARI and mBO---DINOSAUR + SLP reliably outperforms DINOSAUR.

\begin{table} [t]
\caption{Real-World Image Experiments}
\label{tab:boqsareal}
\centering
\begin{subtable}[t]{0.99\textwidth}
\caption{ Unsupervised foreground extraction performance on BO-QSA \citep{jia2022unsupervised} with and without SLP. Results from earlier baseline models are based on the \emph{best} replication, whereas the our results are \emph{mean} ~± 1 SEM performance across three replications per simulation.
}
\centering
\resizebox{\columnwidth}{!}{%
\begin{tabular}{ lllllll} 
\toprule
Method & \multicolumn{2}{c}{\textbf{CUB}} & \multicolumn{2}{c}{\textbf{Stanford Dogs}} & \multicolumn{2}{c}{\textbf{Stanford Cars}} \\
\cmidrule(r){2-3} \cmidrule(r){4-5} \cmidrule(r){6-7}
& IoU & Dice & IoU & Dice & IoU & Dice \\
\midrule
ReDO \citep{Chen2019UnsupervisedOS} &
0.46 &
0.60 &
0.55 &
0.70 &
0.52 &
0.68 \\

IODINE \citep{IODINE} &
0.30 &
0.44 &
0.54 &
0.67 &
0.51 &
0.67 \\

OneGAN \citep{Benny2019OneGANSU} &
0.55 &
0.69 &
0.71 &
0.81 &
0.71 &
0.82 \\

SLATE \citep{Singh2022SLATE} &
0.36 &
0.51 &
0.62 &
0.76 &
0.75 &
0.85 \\

I-SA \citep{Chang2022ObjectRA} &
0.63 &
0.72 &
\cellcolor{blue!15}0.80 &
\cellcolor{blue!15}0.89 &
\cellcolor{blue!15}0.85 &
\cellcolor{blue!15}0.92 \\

BO-QSA \citep{jia2022unsupervised} & 
0.61 \textcolor{gray}{{\ssmall ~± 0.12}} &  %cub iou
0.74 \textcolor{gray}{{\ssmall ~± 0.12}} & %cub dice
0.78 \textcolor{gray}{{\ssmall ~± 0.12}} & % sd iou
0.68 \textcolor{gray}{{\ssmall ~± 0.04}} & % sd dice
0.76 \textcolor{gray}{{\ssmall ~± 0.05}} & % cars iou
0.86 \textcolor{gray}{{\ssmall ~± 0.05}} \\  % cars dice

\midrule

BO-QSA + SLP & 
\cellcolor{blue!15}0.68 \textcolor{gray}{\ssmall ~± 0.02} & % cub iou
\cellcolor{blue!15}0.80 \textcolor{gray}{\ssmall ~± 0.02} & % cub dice
0.78 \textcolor{gray}{\ssmall ~± 0.12} & % sd iou
0.87 \textcolor{gray}{\ssmall ~± 0.12} & % sd dice
0.82 \textcolor{gray}{\ssmall ~± 0.08} & %cars iou
0.91 \textcolor{gray}{\ssmall ~± 0.05} \\ % cars dice

\bottomrule
\end{tabular} %
}

\label{tab:boqsareal1}
% \vspace{-0.1cm}
\end{subtable}
\hfill
\begin{subtable}[t]{0.99\textwidth}
\caption{Unsupervised multi-object segmentation using BO-QSA with and without SLP,\\ following \cite{yang2022} (mean ~± 1 SEM across 3 replications of each simulation)\\ }
\centering
\resizebox{\columnwidth}{!}{%
\begin{tabular}{ lcccccccc} 
\toprule
Method & \multicolumn{4}{c}{\textbf{COCO}} & \multicolumn{4}{c}{\textbf{ScanNet}} \\
\cmidrule(r){2-5} \cmidrule(r){6-9}
& AP@05 & Precision & Recall & PQ & AP@05 & Precision & Recall & PQ \\
\midrule
BO-QSA &
0.093 \textcolor{gray}{\ssmall ~± 0.152} &
0.169 \textcolor{gray}{\ssmall ~± 0.194} &
0.208 \textcolor{gray}{\ssmall ~± 0.210} &
0.114 \textcolor{gray}{\ssmall ~± 0.159} &
0.234 \textcolor{gray}{\ssmall ~± 0.057} &
0.339 \textcolor{gray}{\ssmall ~± 0.054} &
0.387 \textcolor{gray}{\ssmall ~± 0.060} &
0.234 \textcolor{gray}{\ssmall ~± 0.051} \\

BO-QSA + SLP & 
\cellcolor{blue!15}0.126 \textcolor{gray}{\ssmall ~± 0.115} &
\cellcolor{blue!15}0.206 \textcolor{gray}{\ssmall ~± 0.129} &
\cellcolor{blue!15}0.251 \textcolor{gray}{\ssmall ~± 0.142} &
\cellcolor{blue!15}0.139 \textcolor{gray}{\ssmall ~± 0.100} &
\cellcolor{blue!15}0.261 \textcolor{gray}{\ssmall ~± 0.060} &
\cellcolor{blue!15}0.349 \textcolor{gray}{\ssmall ~± 0.048} &
\cellcolor{blue!15}0.399 \textcolor{gray}{\ssmall ~± 0.051} &
\cellcolor{blue!15}0.242 \textcolor{gray}{\ssmall ~± 0.040} 
\\
\bottomrule
\end{tabular} %
}

\label{tab:boqsareal2}
%\vspace{-0.3cm}
\end{subtable}
\end{table}

\begin{figure}[b]
    \centering
    \includegraphics[width=\linewidth]{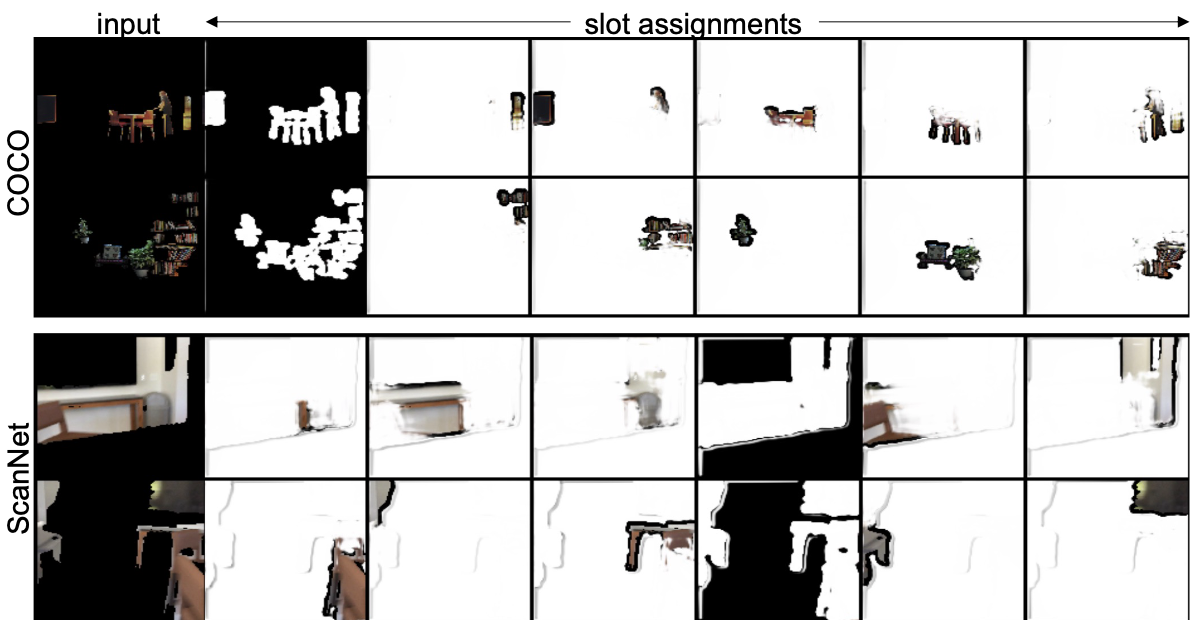}
    \caption{Slot assignments for BO-QSA + SLP on COCO and ScanNet.}
    \label{fig:rw_viz}
    %\vspace{-0.3cm}
\end{figure}

\subsection{Object Discovery in Real-World Images}
\label{ssec:rwexpts}
For real-world data, following \cite{jia2022unsupervised} we use two tasks to evaluate BO-QSA + SLP: unsupervised foreground extraction and unsupervised multi-object segmentation.  For unsupervised foreground extraction, we experiment on the CUB \citep{Wah2011TheCB}, Stanford Dogs \citep{Khosla2012NovelDF}, and Stanford Cars \citep{Krause20133DOR} datasets and evaluate using Intersection-over-Union (IoU) and Dice evaluation metrics. For unsupervised multi-object segmentation, we experiment on COCO \citep{cocodataset} and ScanNet \citep{scannet} datasets and evaluate using the metrics followed in \cite{yang2022} 

\textit{Methodology.} 
As our base model, we use BO-QSA \citep{jia2022unsupervised} with the SLATE encoder-decoder setup \citep{Singh2022SLATE}, which consists of a $4$-layer CNN encoder and a Transformer decoder in a dVAE setup. The slots are initialized as learned embeddings and the initializations are optimized directly as in \citet{jia2022unsupervised}. During evaluation, we select predicted foreground as the one with the maximum intersection between slot's mask prediction and ground-truth foreground mask. 

\textit{Results.} Table \ref{tab:boqsareal1} presents results for unsupervised foreground extraction and Table \ref{tab:boqsareal2} presents results for unsupervised multi-object segmentation. On all data sets and on both tasks, SLP consistently improves the performance of BO-QSA, the most significant improvement being on the Dice metric for the Stanford Dogs dataset.
It appears that I-SA \citep{Chang2022ObjectRA} marginally outperforms BO-QSA + SLP on Stanford Dogs and Stanford Cars. \emph{However},
the baselines reported in Table~\ref{tab:boqsareal1}---including I-SA results---are the best performing replication of multiple runs, whereas we report mean performance across three replications for BO-QSA and BO-QSA + SLP.
Figure \ref{fig:rw_viz} presents visualizations of slot assignments for several
COCO images (first and second rows) and ScanNet images (third and fourth rows).

\begin{table} [t]
\caption{FG-ARI (\%) and Mean-Squared Error (MSE)\\
BO-QSA \cite{jia2022unsupervised} with and without SLP \\(mean ~± 1 SEM across 3 experiment trials)}
\centering

\begin{tabular}{ lcc} 
\toprule
Method & \multicolumn{2}{c}{\textbf{ClevrTex-OOD}} \\
\cmidrule(r){2-3}
& FG-ARI & MSE  \\
\midrule
MONet \citep{burgess2019monet} & 
0.37 \textcolor{gray}{{\ssmall ~± 0.05}} &
409 \textcolor{gray}{{\ssmall ~± 0.5}} \\

GENESIS-V2 \citep{Engelcke2021GENESISV2IU} & 
0.29 \textcolor{gray}{{\ssmall ~± 0.19}} &
539 \textcolor{gray}{{\ssmall ~± 7.0}} \\

Slot-Attention \citep{Locatello2020} & 
0.58 \textcolor{gray}{{\ssmall ~± 0.05}} & %or
487 \textcolor{gray}{{\ssmall ~± 2.3}} \\

I-SA \citep{Chang2022ObjectRA} &
0.83 \textcolor{gray}{{\ssmall ~± 0.05}} &
241 \textcolor{gray}{{\ssmall ~± 1.1}} \\

BO-QSA \citep{jia2022unsupervised} & 
0.86 \textcolor{gray}{{\ssmall ~± 0.05}} & %or
265 \textcolor{gray}{{\ssmall ~± 2.9}} \\ %ct

\midrule

BO-QSA + SLP & 
\cellcolor{blue!15}0.88 \textcolor{gray}{{\ssmall ~± 0.05}} & %or
\cellcolor{blue!15}243 \textcolor{gray}{{\ssmall ~± 1.6}} \\ %ct

%BO-QSA + SLP (20 spatial iterations) & 
%\cellcolor{blue!15}0.89 \textcolor{gray}{{\ssmall ~± 0.05}} & %or
%\cellcolor{blue!15}239 \textcolor{gray}{{\ssmall ~± 1.4}} \\ %ct
%
%BO-QSA + SLP (25 spatial iterations) & 
%\cellcolor{blue!15}0.89 \textcolor{gray}{{\ssmall ~± 0.05}} & %or
%\cellcolor{blue!15}242 \textcolor{gray}{{\ssmall ~± 0.8}} \\ %ct

\bottomrule
\end{tabular} 
\label{tab:boqsaood}
%\vspace{-0.3cm}
\end{table}

\subsection{Out-of-Distribution Generalization}
One potential weakness of our model is the explicit learning of $\bm{\alpha}^0$ based on training-set-wide statistics of the likely spatial locations of objects. This initialization may result in poor OOD generalization. However, we observe that this issue does not adversely affect our model. As suggested by Figure \ref{fig:alpha_init}, learning $\bm{\alpha}^0$ does help break symmetry: the learned $\bm{\alpha}$ disperses the slot means across the scene while maintaining a high slot variance. As a result, the attention distribution forces the slots to model different objects in the scene. However, since the learned initialization is normalized before further optimization, the effects of an excessively large $\bm{\alpha}^0$ initialization should be effectively nullified. 

As evidence in support of our conjecture, we tested Slot Attention with SLP and initialized slots as learned embeddings as in Section \ref{ssec:bqsa}. We evaluate object discovery on the ClevrTex-OOD split variants and present results in Table \ref{tab:boqsaood}. Empirically, as we see improvements both in FG-ARI and in MSE Reconstruction loss, we are able to prove that learning dataset wide statistics merely has the effect of learning an $\bm{\alpha}^0$ initialization that supports symmetry breaking between slots.

\section{Discussion}

In this paper, we proposed a spatial-locality prior that is consistent with both human visual attention and statistics of objects in natural images. Incorporating this prior into unsupervised object-centric models  biases slot decompositions of the visual scene when direct evidence for objects and their boundaries is weak. The result is improved models that are more robust to hyperparameter selection and that yield better object segmentations. We show consistent improvements with three object-centric architectures (Slot Attention, BO-QSA, and DINOSAUR), eight distinct data sets,  and various performance measures that have been used in the literature, including FG-ARI, mBO, IoU, and Dice. In all cases, models incorporting SLP advance state-of-the-art performance. 

%In this paper, we present Spotlight Attention, a method which is capable of unsupervised object-centric learning on both synthetic and real world datasets. This is achieved through addressing the inadequate spatial bias present in Slot Attention and its variants. We show that our model is invariant to other architectural considerations and brings significant improvements to various different baselines agnostic of dataset. 

\textit{Limitations and Future Work.}  A key limitation of the proposed method---as well as of any slot-based object-centric model, is the hard requirement to specify the maximum number of slots that the model can represent. Another limitation of SLP in particular is the increased computational complexity due to the bi-level optimization algorithm. Fortunately, we have found that a single spatial iteration, which is relatively efficient, yields significant benefits. In future we hope to extend the proposed method from static images to video streams. SLP should be even more fruitful for video streams where the spatial modulations, $\bm{\alpha}$, inferred for one frame should be a suitable initialization point for the next frame. We also hope to extend the method to include the depth dimension of spatial attention, allowing SLP to operate in depth and to thereby predict occlusions.

%, the proposed method can be easily ported to 3D images and effectively use depth in the $\bm{\alpha}$ optimization procedure.

\section{Acknowledgement}
This research was enabled in part by compute resources provided by Mila (mila.quebec). We would like to thank Vedant Shah and Aniket Didolkar for reviewing early versions of the manuscript. We would also like to thank Mihir Prabhudesai and Katerina Fragkiadaki for useful discussions.

\bibliographystyle{plainnat}
\bibliography{main}

\newpage

\section{SLP Configuration}

\begin{table}[t]
    \caption{Spatial Locality Prior Configuration Sweep}
    \centering
    \begin{tabular}{cc}
      \toprule
      $\bm{\alpha}_{\textrm{lr}}$  & [1.0, 0.5, 0.1] \\
      $\lambda_{norm}$ & 0.1  \\
      spatial iterations & [1, 5, 10, 20, 25] \\ 
      \bottomrule
    \end{tabular}
    \label{tab:spatialconfig}
\end{table}

In order to ensure robust experimentation, we adhere to the default hyperparameters\footnote{For details on the default hyperparameters, please refer to the original papers.} used in the previously reported results. We achieve this by augmenting various open-sourced and official implementations of each model, which serve as our baselines. The only exception we made was in determining the batch size, which was influenced by the hardware available to us.

Since each dataset and model exhibit different learning dynamics and scene complexities, we tune the hyperparameters of the SLP according to the sweep configuration outlined in Table \ref{tab:spatialconfig}. Through experimentation, we have found that setting $\bm{\alpha}_{\textrm{lr}}$ to $1.0$ consistently yields the best results. Consequently, we exclude the inferior results obtained with other values of $\bm{\alpha}_{\textrm{lr}}$ from our analysis.

Generally, we observe that a larger number of spatial iterations is necessary to learn the appropriate spatial bias for scenes derived from more complex real-world datasets like COCO and ScanNet. Conversely, smaller values of spatial iterations tend to suffice for simpler and synthetic datasets such as ClevrTex. For example, as shown in Table \ref{tab:ali-ablate}, setting $T_{spat} = 1$ brings about a significant performance improvement for 7-slot Slot Attention on ClevrTex.

Furthermore, we have discovered that annealing $\bm{\alpha}_{\textrm{lr}}$ by a factor of $(T_{spat} - j) \ / \ T_{spat}$ enhances optimization dynamics. Our hypothesis is that initializing $\bm{\alpha}_{\textrm{lr}}$ at $1.0$ and subsequently annealing it results in a few aggressive steps towards the optimum, followed by several fine-tuning steps.

\subsection{Vanilla Slot Attention Experiments}
For the experiments conducted in Section \ref{sssec:sa}, we utilized the open-source implementation provided by \cite{pmlr-v162-dittadi22a}\footnote{https://github.com/addtt/object-centric-library}. In Table \ref{tab:spatialsaspatial}, an extension of Table \ref{tab:ali-ablate}, we include higher values of spatial iterations. However, we observe minimal gains in performance when increasing the spatial iterations from 5 to 15. This leads us to hypothesize that the design of annealing $\bm{\alpha}_{\textrm{lr}}$ starting from $1.0$ encourages rapid convergence to a near-optimal $\bm{\alpha}$, and further increasing the number of spatial iterations does not yield significant performance improvements. Importantly, we did not observe any degradation in performance for any number of spatial iterations, indicating that SLP does not introduce additional gradient instability as observed when increasing the number of slot iterations in Slot Attention and its variants.

\begin{table}[h]
    \caption{Foreground ARI (\%) Segmentation Accuracy \\ (mean ~± 1 SEM across 3 replications of each simulation) \\
    Varying spatial iterations for 7-slot Slot Attention on ClevrTex \\}
    \centering
    \begin{tabular}{lc}
        \toprule
        Slot Attention \citep{Locatello2020} & 
        0.47 \textcolor{gray}{{\ssmall ~± 0.22}} \\ %10

        \midrule

        Slot Attention + SLP (1 spatial iteration) & 
        0.54 \textcolor{gray}{\ssmall ~± 0.05} \\ % 10

        Slot Attention + SLP (5 spatial iterations) & 
        0.64 \textcolor{gray}{\ssmall ~± 0.10} \\ % 1

        Slot Attention + SLP  (10 spatial iterations) & 
        0.65 \textcolor{gray}{\ssmall ~± 0.08} \\ % 10

        Slot Attention + SLP  (15 spatial iterations) & 
        0.66 \textcolor{gray}{\ssmall ~± 0.05} \\ 

        \bottomrule
    \end{tabular}
    \label{tab:spatialsaspatial}
\end{table}

\subsection{BO-QSA experiments}
For the experiments conducted in Sections \ref{ssec:bqsa} and \ref{ssec:rwexpts}, we built on the open-source implementation provided by \cite{jia2022unsupervised}\footnote{https://github.com/YuLiu-LY/BO-QSA}. In Tables \ref{tab:boqsasynthspatial} and \ref{tab:boqsaoodspatial}, we extend Tables \ref{tab:boqsasynth} and \ref{tab:boqsaood} respectively to include higher values of spatial iterations. Through these extensions, we were able to validate our hypothesis regarding the lack of improvement beyond a certain number of spatial iterations.

Specifically, in Table \ref{tab:boqsasynth}, we observe that increasing the number of spatial iterations to 5 does not result in a significant performance improvement on ClevrTex. However, as demonstrated in Table \ref{tab:spatialsaspatial}, setting SLP to 5 spatial iterations leads to a significant enhancement in performance. This suggests that the hyperparameter for spatial iterations is sensitive not only to dataset complexity but also to the learning dynamics of the model being used.

\begin{table} [h]
\caption{Foreground ARI (\%) Segmentation Accuracy \\ (mean ~± 1 SEM across 3 replications of each simulation)\\
BO-QSA \citep{jia2022unsupervised} augmented with SLP \\
Varying spatial iterations \\}
\centering

\begin{tabular}{ lcccccc} 
\toprule
Method & \textbf{ShapeStacks} &\textbf{ObjectsRoom} & \textbf{ClevrTex} \\
\midrule
BO-QSA \citep{jia2022unsupervised} & 
0.93 \textcolor{gray}{{\ssmall ~± 0.05}} &  %s
0.87 \textcolor{gray}{{\ssmall ~± 0.05}} & %or
0.80 \textcolor{gray}{{\ssmall ~± 0.08}} \\ %ct

\midrule

BO-QSA + SLP (5 spatial iterations) & 
0.93 \textcolor{gray}{\ssmall ~± 0.10} & % ss
0.92 \textcolor{gray}{\ssmall ~± 0.08}  & % or
0.82 \textcolor{gray}{\ssmall ~± 0.05}  \\ % ct

BO-QSA + SLP (10 spatial iterations) & 
0.95 \textcolor{gray}{\ssmall ~± 0.08} &   % ss
0.93 \textcolor{gray}{\ssmall ~± 0.05} & % or
0.87 \textcolor{gray}{\ssmall ~± 0.05} \\ % ct

\bottomrule
\end{tabular} 
\label{tab:boqsasynthspatial}
% \vspace{-0.1cm}
\end{table}

\begin{table} [h]
\caption{FG-ARI (\%) and Mean-Squared Error (MSE)\\
BO-QSA \cite{jia2022unsupervised} with and without SLP \\(mean ~± 1 SEM across 3 experiment trials) \\ 
Varying spatial iterations \\}
\centering

\begin{tabular}{ lcc} 
\toprule
Method & \multicolumn{2}{c}{\textbf{ClevrTex-OOD}} \\
\cmidrule(r){2-3}
& FG-ARI & MSE  \\
\midrule

BO-QSA \citep{jia2022unsupervised} & 
0.86 \textcolor{gray}{{\ssmall ~± 0.05}} & %or
265 \textcolor{gray}{{\ssmall ~± 2.9}} \\ %ct

\midrule

BO-QSA + SLP (10 spatial iterations) & 
0.88 \textcolor{gray}{{\ssmall ~± 0.05}} & %or
243 \textcolor{gray}{{\ssmall ~± 1.6}} \\ %ct

BO-QSA + SLP (20 spatial iterations) & 
0.89 \textcolor{gray}{{\ssmall ~± 0.05}} & %or
239 \textcolor{gray}{{\ssmall ~± 1.4}} \\ %ct
BO-QSA + SLP (25 spatial iterations) & 
0.89 \textcolor{gray}{{\ssmall ~± 0.05}} & %or
242 \textcolor{gray}{{\ssmall ~± 0.8}} \\ %ct

\bottomrule
\end{tabular} 
\label{tab:boqsaoodspatial}
% \vspace{-0.1cm}
\end{table}

\subsection{DINOSAUR experiments}
For the experiments conducted in Section \ref{sssec:dinosaur}, we built upon the open-source implementation provided by \cite{Seitzer2023}\footnote{https://github.com/amazon-science/object-centric-learning-framework}. In these experiments, we were able to run higher values of spatial iterations, as demonstrated in Table \ref{tab:dinosaursynthspatial}. This was possible because DINOSAUR utilizes a frozen pre-trained ViT-B/8 encoder, resulting in a lower number of effective model parameters. Interestingly, we observed significant improvements for higher values of spatial iterations, which can be attributed to the fact that the DINOSAUR model does not apply any positional encoding to the encoded image representation. These results provide further evidence that SLP offers a stronger spatial bias compared to the positional encoding layer. They also suggest that existing methods which use positional encoding layers, need a stronger spatial bias which SLP provides. 

\begin{table}[h]
%\begin{wraptable}{r}{8.5cm}
%\tiny
    %\vspace{-0.5cm}
    \centering
    \caption{FG-ARI and mBO measures of object discovery\\ on MoVi-C and MoVi-E with DINOSAUR \citep{Seitzer2023} (mean ~± 1 SEM across 3 runs).\\}

    \begin{tabular}{ lcccc} 
    \toprule
    Method & \multicolumn{2}{c}{\textbf{MoVi-C}} & \multicolumn{2}{c}{\textbf{MoVi-E}} \\
\cmidrule(r){2-3}   \cmidrule(r){4-5}
    & FG-ARI & mBO & FG-ARI & mBO\\
    \midrule
    DINOSAUR (ViT-B/8) &
    68.9 \textcolor{gray}{\tiny ~± 0.3} &
    38.0 \textcolor{gray}{\tiny ~± 0.2} &
    65.1 \textcolor{gray}{\tiny ~± 0.6} &
    33.5 \textcolor{gray}{\tiny ~± 0.1} \\

    \midrule

    DINOSAUR + SLP (10 spatial iterations) & 
     70.1 \textcolor{gray}{\tiny ~± 0.4} &
     38.5 \textcolor{gray}{\tiny ~± 0.2} &
     66.3 \textcolor{gray}{\tiny ~± 0.6} &
     34.5 \textcolor{gray}{\tiny ~± 0.3} \\

    DINOSAUR + SLP (20 spatial iterations) & 
    \cellcolor{blue!15}72.8 \textcolor{gray}{\tiny ~± 0.6} &
    \cellcolor{blue!15}41.5 \textcolor{gray}{\tiny ~± 0.3} &
    \cellcolor{blue!15}70.4 \textcolor{gray}{\tiny ~± 0.4} &
    \cellcolor{blue!15}35.9 \textcolor{gray}{\tiny ~± 0.2} \\

    DINOSAUR + SLP (50 spatial iterations) & 
    74.6  &
    42.6  &
    73.1  &
    36.7  \\

    \bottomrule
    \end{tabular} 
    \label{tab:dinosaursynthspatial}
    %\vspace{0.1cm}
%\end{wraptable}
\end{table}

\end{document}